\documentclass[conference]{IEEEtran}
\IEEEoverridecommandlockouts

\usepackage{cite}
\usepackage{tabularx}
\usepackage{amsmath,amssymb,amsfonts}
\usepackage{algorithmic}
\usepackage{graphicx}
\usepackage{textcomp}
\usepackage{xcolor}
\usepackage{cite}
\usepackage{booktabs}
\usepackage{tabularx}
\usepackage{array}
\usepackage{multirow}
\usepackage{multicol}

\newcolumntype{Y}{>{\centering\arraybackslash}X}
\usepackage[hidelinks]{hyperref}

\def\BibTeX{{\rm B\kern-.05em{\sc i\kern-.025em b}\kern-.08em
    T\kern-.1667em\lower.7ex\hbox{E}\kern-.125emX}}
\begin{document}

\title{Understanding Multilingual Medical ASR Adaptation Through Layer-Wise Analysis
{\footnotesize \textsuperscript{}}
\thanks{}
}

\author{
\IEEEauthorblockN{
Souranil Kahali\textsuperscript{1$\star$},
Rituparna Bose\textsuperscript{1$\star$},
Abner Hernandez\textsuperscript{1},
Tomas Arias-Vergara\textsuperscript{1},
Andreas Maier\textsuperscript{1},\\
Ning Ma\textsuperscript{2},
Paula A. Perez-Toro\textsuperscript{1,3}
}
\\
\IEEEauthorblockA{
\textsuperscript{1}Pattern Recognition Lab, Friedrich-Alexander-Universität Erlangen-Nürnberg (FAU), Erlangen, Germany\\
\textsuperscript{2}School of Computer Science, University of Sheffield, Sheffield, UK\\
\textsuperscript{3}Chair for AI in Healthcare and Medicine, Technical University of Munich (TUM), Munich, Germany\\
\textsuperscript{$\star$}Equal contribution\\
}
}

\maketitle
\begin{abstract}

Medical automatic speech recognition (MedASR) requires adaptation to specialised terminology, limited annotated clinical data, and multilingual use cases. Although large-scale pretrained ASR models such as Whisper achieve strong generalisation, their behaviour after medical and multilingual adaptation remains insufficiently understood beyond word error rate (WER). This paper investigates how multilingual medical adaptation reshapes the internal representations of Whisper models through layer-wise encoder analysis. We compare zero-shot decoding, English-only fine-tuning, German-only diagnostic fine-tuning, two-stage EN$\rightarrow$EN+DE continuation, and direct EN+DE fine-tuning across Whisper model sizes. Fine-tuning substantially improves MedASR performance, but the best model depends on the adaptation setting: Whisper-Medium gives the lowest English WER (7.72\%) and the lowest combined EN+DE WER under direct EN+DE training (26.30\%); German-only Whisper-Large-v3 gives the lowest German WER (44.96\%), but as a within-corpus diagnostic on 86 single-speaker training utterances rather than robust generalisation. Layer-wise analysis of the two-stage Whisper-Small trajectory shows that English medical fine-tuning produces the dominant encoder shift, whereas multilingual continuation largely preserves the adapted representation space. Domain and language information remain highly recoverable across layers, while linearly recoverable error-predictive cues weaken as WER improves.

\end{abstract}


\section{Introduction}
Automatic speech recognition (ASR) systems~\cite{6638947,10.1109/ICASSP.2016.7472621,zhang2020transformer} have made substantial gains in recent years, through Transformer-based architectures and large-scale pretraining. Models such as Whisper~\cite{radford2023robust} and Wav2Vec2~\cite{baevski2020wav2vec} achieve strong performance on general-purpose benchmarks like LibriSpeech~\cite{panayotov2015librispeech} and CommonVoice~\cite{ardila2020commonvoicemassivelymultilingualspeech}. However, adapting these systems to domain-specific applications-particularly healthcare-presents distinct challenges. Medical speech contains rare and specialised terminology, recorded under varied speaker and acoustic conditions, and high accuracy demands where a single transcription error can carry clinical consequences. Generic ASR models trained on general speech corpora perform poorly in these settings, which motivates targeted domain adaptation.

Prior work has shown that fine-tuning pretrained medical ASR (MedASR) models on domain-specific corpora substantially improves recognition accuracy~\cite{adedeji2024sound,hernandez2025enhancing}. Multilingual MedASR has also been explored through initiatives such as MultiMed~\cite{le2025multimed}, which introduced benchmarks across clinical languages. In parallel, interpretability research has established that Transformer-based speech encoders organise information hierarchically: lower layers tend to encode acoustic and phonetic information, while deeper layers capture abstract linguistic representations~\cite{pasad2021layer,belinkov-glass-2019-analysis}. However, these two research directions, domain adaptation and representation analysis, have largely developed independently. In particular, it remains unclear how medical and multilingual fine-tuning reorganise the internal encoder representations of pretrained ASR models.

Our work aims to close that gap by combining multi-size Whisper evaluation with systematic layer-wise analysis of how medical and multilingual adaptation reshape encoder representations. We define the two-stage EN$\rightarrow$EN+DE setting used throughout the paper as English medical fine-tuning followed by continued fine-tuning on combined English–German medical speech initialised from the English checkpoint. The main contributions are: (i) A systematic comparison of zero-shot, monolingual, continued EN$\rightarrow$EN+DE, and direct EN+DE Whisper adaptation across four model sizes; (ii) A layer-wise analysis showing that English medical fine-tuning produces the dominant representational change, whereas multilingual continuation largely preserves the English-adapted representations; and (iii) Evidence that encoder representations become less linearly predictive of transcription errors following adaptation, even as decoder WER improves. 
Fig.~\ref{fig:overview} provides a visual overview of the study design. It summarises the data and adaptation pipeline, the main ASR results, and the layer-wise analyses used to interpret the adapted encoder representations.

\begin{figure*}[!htpb]
    \centering
    \includegraphics[width=0.9\textwidth]{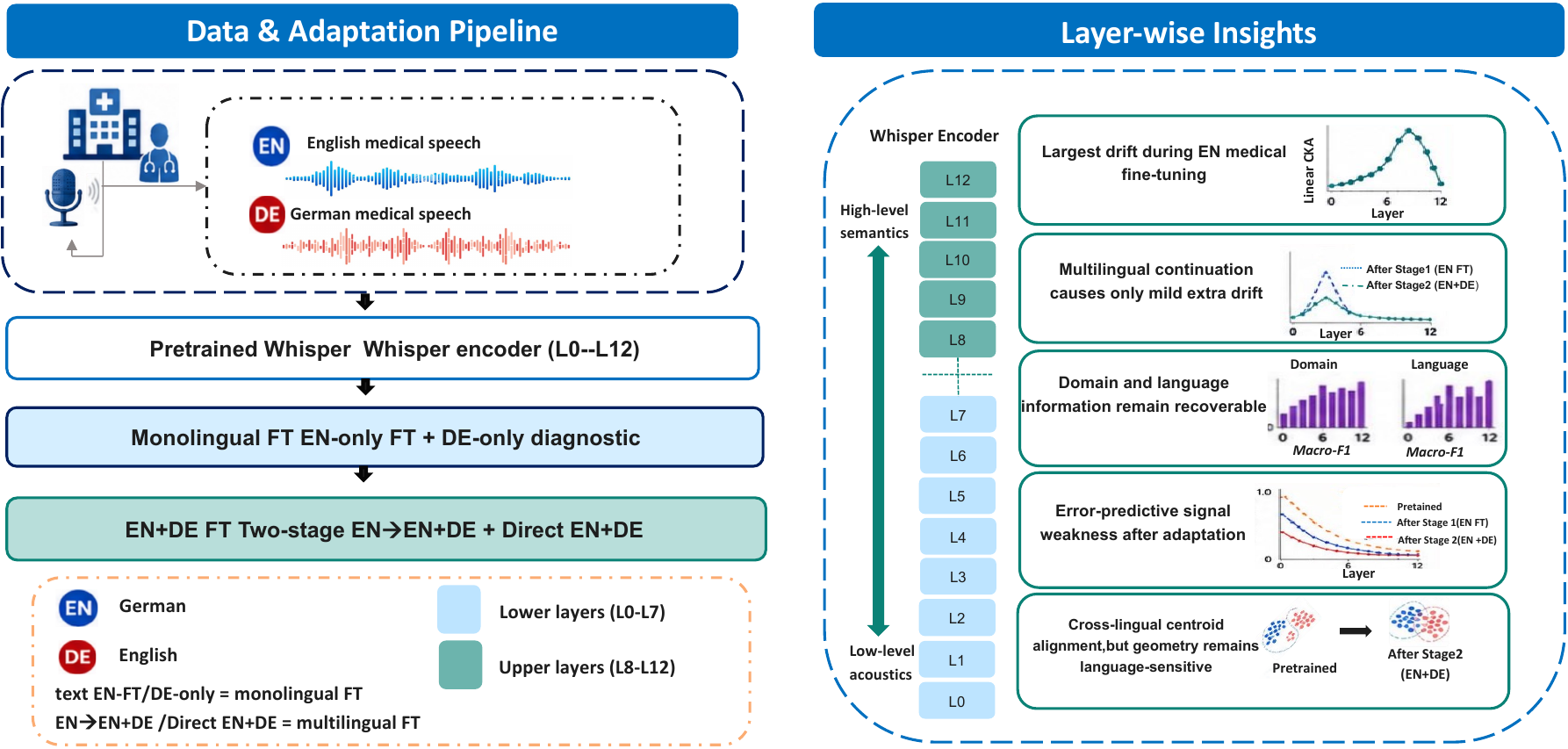}
\caption{Overview of the multilingual MedASR adaptation and analysis pipeline. English and German medical speech are used for monolingual fine-tuning, two-stage EN→EN+DE continuation, and direct EN+DE fine-tuning. Adapted models are evaluated with WER, and Whisper-Small encoder representations are analysed through drift and probing tasks.}
\label{fig:overview}
    \label{fig:overview}
\end{figure*}

\section{Related Work}

\textbf{Transformer-based ASR.}
Large-scale pretrained ASR models are strong starting points for domain adaptation, but their WER performance depends heavily on the target domain. Whisper~\cite{radford2023robust} is an encoder--decoder Transformer trained on large-scale weakly supervised multilingual audio, giving it strong zero-shot and cross-lingual ASR capabilities. Wav2Vec2~\cite{baevski2020wav2vec} uses self-supervised contrastive pretraining and achieves low WER on standard benchmarks after supervised fine-tuning. However, strong general-domain performance does not necessarily transfer to medical speech, where specialised terminology and variable acoustic conditions require targeted adaptation~\cite{adedeji2024sound,liu2024exploration}.

\textbf{Medical ASR.}
Fine-tuning pretrained ASR models on clinical corpora has consistently improved medical transcription performance~\cite{adedeji2024sound,le2025multimed}. Adedeji et al.~\cite{adedeji2024sound} showed that domain-specific fine-tuning, combined with large language model  post-processing, improves medical ASR accuracy. The MultiMed project~\cite{le2025multimed} introduced multilingual medical ASR across five languages, demonstrating the feasibility of adapting general ASR models to clinical speech. Its results are not directly comparable to ours because the datasets, languages, clinical scenarios, train/test splits, and decoding setup differ from the English Kaggle and German PoCaP evaluation used here. We therefore restrict direct empirical comparisons to models evaluated on the same test utterances under the same decoding and metric pipeline. Our work builds on this line by studying English--German medical adaptation across multiple Whisper fine-tuning strategies and by providing layer-level evidence of how adaptation changes encoder representations.

\textbf{Probing and layer-wise analysis.}
Belinkov and Glass~\cite{belinkov-glass-2019-analysis} surveyed probing methods for analysing neural language models, establishing the use of lightweight classifiers on frozen representations. Pasad et al.~\cite{pasad2021layer} applied layer-wise analysis to self-supervised speech models, showing that different encoder depths encode different phonological and linguistic properties. Prasad and Jyothi~\cite{prasad2020accents} probed accent information in end-to-end ASR systems. Wiepert et al.~\cite{wiepert2024speech} examined layer selection for pathological speech feature prediction. To our knowledge, no prior work has jointly examined two-stage multilingual medical ASR adaptation and layer-wise probing of domain, language, and error-predictability signals in a fine-tuned encoder--decoder ASR model.

\textbf{Clinical speech analysis.} Prior work on pathological speech has often combined acoustic and linguistic embeddings within Transformer-based pipelines~\cite{wiepert2024speech,perez2021influence}. These approaches focus on supervised classification tasks. Our work takes a different angle: we analyse the encoder's layer-wise behaviour for interpretability, without relying on diagnostic labels.

\section{Data and Pre-Processing}

This section describes the medical speech datasets and pre-processing pipeline used for the monolingual and multilingual Whisper fine-tuning experiments.

\textbf{English Medical Corpus.}
We use the publicly available \textit{Kaggle Medical Speech, Transcription, and Intent} dataset \cite{kaggle_medical_speech}, comprising approximately 8.5 hours of spoken medical phrases with human-written transcriptions. Audio files are resampled to 16~kHz mono before fine-tuning.
The dataset is partitioned into speaker-disjoint training, validation, and test splits using speaker identifiers: training (997 samples, 23 speakers), validation (136 samples, 3 speakers), and test (104 samples, 3 speakers). No speaker appears in more than one split, enabling speaker-disjoint evaluation. Mean transcript length is 49.4 characters in training and 52.1 characters in testing.

\textbf{German Medical Corpus.}
German data are drawn from the PoCaP Corpus~\cite{demir2022pocap}, which contains port catheter placement procedures recorded at the Radiology Department of University Hospital Erlangen, Germany. The dataset is not publicly available due to patient privacy regulations. We use audio from the operating physician only. Training uses operation OP-001 (86 samples), while validation (37 samples) and test (38 samples) are file-disjoint halves of OP-002, ensuring no audio overlap. The same resampling and normalisation pipeline is applied as for the English corpus. Because the German training set is small and single-speaker, German results reflect within-corpus adaptation rather than cross-speaker generalisation. We therefore report German-only fine-tuning only as a diagnostic baseline, while the main multilingual comparisons use combined EN+DE training.

\textbf{Multilingual Dataset.}
The multilingual dataset is created by merging the English and German splits while preserving text labels and language identifiers. It contains 1,083 training, 173 validation, and 142 test samples, including 104 English and 38 German test utterances. During evaluation, each sample is decoded using language-specific forced decoder identifiers.

\textbf{Pre-processing and normalisation.} Audio is converted to mono and resampled to 16~kHz. We use the corresponding Whisper feature extractor~\cite{radford2023robust, wolf2020huggingfacestransformersstateoftheartnatural} for log-Mel features (80-bin for Base/Small/Medium, 128-bin for Large-v3). Transcripts are kept in their original casing and punctuation; no text normalisation, denoising, voice-activity detection, or augmentation is applied.

\textbf{General English Contrast Set}
For the domain probing task, a 100-sample subset of LibriSpeech test-clean~\cite{panayotov2015librispeech} is used as a general English contrast. Selecting samples with the same language as the medical set controls for language effects, ensuring that domain probe performance reflects domain-specific encoder content rather than language discrimination.

\textbf{Reproducibility.} We will release the English split manifests, preprocessing scripts, evaluation code, and experiment configurations. The German PoCaP audio cannot be redistributed due to privacy restrictions, but split identifiers and evaluation scripts will be provided.

\section{Fine-Tuning Methodology}
\label{sec:fine-tuning}
\subsection{Training Configuration}
All fine-tuning experiments are implemented using Hugging Face Transformers~\cite{wolf2020huggingfacestransformersstateoftheartnatural} and PyTorch~\cite{paszke2019pytorchimperativestylehighperformance} on a single NVIDIA A100 GPU. We evaluate four fine-tuning settings: English-only medical fine-tuning, German-only diagnostic fine-tuning, two-stage EN$\rightarrow$EN+DE continuation, and direct EN+DE fine-tuning from pretrained weights.

Training uses Seq2SeqTrainer with cross-entropy loss, greedy decoding for validation WER, and a maximum generation length of 225 tokens. All runs use an effective batch size of 32, mixed precision, gradient checkpointing, and early stopping with patience 5 based on validation WER. Checkpoints are saved every 200 steps, and the best checkpoint is selected by lowest validation WER.
We compute word error rate (WER), character error rate (CER), and SemScore~\cite{aynetdinov2024semscoreautomatedevaluationinstructiontuned}, where SemScore is the mean cosine similarity between reference and prediction sentence embeddings using \texttt{paraphrase-multilingual-MiniLM-L12-v2}~\cite{reimers2019sentencebertsentenceembeddingsusing}. For validation and final evaluation, Whisper outputs are generated with greedy decoding and language-specific forced decoder identifiers implemented through the Hugging Face Transformers interface~\cite{radford2023robust, wolf2020huggingfacestransformersstateoftheartnatural}. English-only and German-only checkpoints are decoded on the corresponding monolingual test sets, while EN+DE checkpoints are decoded on language-filtered subsets of the multilingual test set and then pooled for the combined score. Wav2Vec2 baselines use CTC argmax decoding\cite{baevski2020wav2vec}. WER and CER are computed with \texttt{jiwer} on the decoded strings without additional case-folding or punctuation removal.
Whisper-Small, Whisper-Medium, and Whisper-Large-v3 are trained under the same configuration; the selected checkpoints are step 400, step 800, and step 400, respectively.

\subsection{Monolingual Medical Fine-Tuning}

For English-only fine-tuning, each Whisper variant is trained on the English medical training set using the shared configuration, with best-checkpoint selection by validation WER. The processor language is fixed to English. The Whisper-Small EN-FT checkpoint used for layer-wise analysis is step 400 (11.98\% English test WER); Whisper-Small is chosen because it balances adapted ASR performance with the compute needed for repeated hidden-state extraction across encoder layers.

The German-only diagnostic follows the same configuration on the 86 German training utterances, with best-checkpoint selection by validation WER. Because this setting uses a small single-speaker German training set, only German test performance is reported and interpreted as within-corpus adaptation evidence.

\subsection{EN+DE Adaptation}
For two-stage adaptation, the English fine-tuned checkpoint is used to initialise continued training on the combined English--German dataset. This setting tests whether English medical adaptation can be preserved while introducing German medical speech. As a second multilingual baseline, direct EN+DE fine-tuning starts from pretrained Whisper weights and trains on the same combined English--German data in one stage. In both settings, no validation or test samples are used for gradient updates, batches mix both languages, and samples are decoded with language-specific forced decoder identifiers during evaluation. For the layer-wise analysis, we use the two-stage Whisper-Small multilingual fine-tuned (ML-FT) checkpoint from step~1,400, which achieves 10.91\% English and 53.87\% German WER.

\section{Layer-Wise Analysis}
We analyse Whisper-Small encoder representations to examine how medical and multilingual fine-tuning affect the internal representation space. The analysis combines hidden-state extraction, representation-drift measurement, and probing classifiers for domain, language, and error-related information.

\subsection{Hidden-State Extraction} 
Whisper-small has 12 encoder Transformer blocks. Including the input embedding output, we extract 13 hidden states per sample (L0--L12). Each hidden state has shape $T \times 384$, where $T$ is the sequence length. We mean-pool each layer over time to obtain one 384-dimensional vector per audio sample and layer. Hidden states are extracted independently from the Pretrained, EN-FT, and multilingual ML-FT checkpoints.

\subsection{Representation Drift} 
Representation drift measures how encoder hidden states change after fine-tuning without using task labels. For the same 100 English audio samples, we compute per-layer similarity between checkpoints using two complementary metrics: cosine similarity between mean representations, which captures directional alignment, and Linear Centered Kernel Alignment (CKA)~\cite{kornblith2019similarity}, which captures changes in representational geometry. We analyse three pairs: Pretrained~$\rightarrow$~EN-FT, EN-FT~$\rightarrow$~ML-FT, and ML-FT English versus ML-FT German.

\subsection{Probing Classifiers.} 
We evaluated three binary probing tasks to analyse what information is encoded across layers.

\noindent\textbf{Domain probe} classifies English medical speech (positive) vs. LibriSpeech~\cite{panayotov2015librispeech} general English speech. This controls for language but may also reflect corpus-specific differences in accent, speaker population, recording conditions, and background noise.

\begin{table}[!htbp]
\centering
\caption{ASR WER (\%) on English and German medical speech test sets. Base (B), Small (S), Medium (M) and Large (L) Whisper versions where adapted.}
\label{tab:asr_main}
\scriptsize
\renewcommand{\arraystretch}{1.05}
\setlength{\tabcolsep}{6pt}
\begin{tabular}{@{}llccc@{}}
\toprule
\textbf{Model} & \textbf{Adapt} &
\textbf{English}$\downarrow$ & \textbf{German}$\downarrow$ & \textbf{Combined}$\downarrow$ \\
\midrule
\multicolumn{5}{@{}c}{\textit{Zero-shot}} \\
\midrule
Whisper$_{B}$ & None & 21.30 & 70.42 & 44.65 \\
Whisper$_{S}$ & None & 17.57 & 62.98 & 39.15 \\
Whisper$_{M}$ & None & 16.77 & 71.99$^\dag$ & 43.02 \\
Whisper$_{L}$ & None & \textbf{16.50} & \textbf{56.51} & \textbf{35.52} \\
\midrule
\multicolumn{5}{@{}c}{\textit{Monolingual (Mono) medical fine-tuning}} \\
\midrule
Whisper$_{B}$ & Mono & 15.00 & 58.86 & -- \\
Whisper$_{S}$ & Mono & 11.98 & 57.69 & -- \\
Whisper$_{M}$ & Mono & \textbf{7.72} & 45.94 & -- \\
Whisper$_{L}$ & Mono & 14.37 & \textbf{44.96} & -- \\
\midrule
\multicolumn{5}{@{}c}{\textit{Two-stage EN$\rightarrow$EN+DE (Two-stage) medical fine-tuning}} \\
\midrule
Whisper$_{B}$ & Two-stage & 14.91 & 58.86 & 35.80 \\
Whisper$_{S}^{\ddag}$ & Two-stage & 10.91 & 53.87 & \textbf{31.33} \\
Whisper$_{M}$ & Two-stage & \textbf{7.99} & 57.49 & 31.52 \\
Whisper$_{L}$ & Two-stage & 13.22 & \textbf{53.09} & 32.17 \\
\midrule
\multicolumn{5}{@{}c}{\textit{Direct multilingual (Multi) medical fine-tuning}} \\
\midrule
Whisper$_{B}$ & Multi & 14.64 & 61.90 & 37.10 \\
Whisper$_{S}$ & Multi & 9.94 & 52.20 & 30.03 \\
Whisper$_{M}$ & Multi & \textbf{7.81} & \textbf{46.72} & \textbf{26.30} \\
Whisper$_{L}$ & Multi & 15.35 & 47.50 & 30.63 \\
\bottomrule
\multicolumn{5}{@{}l@{}}{\scriptsize Mono: EN-only (English) / DE-only diagnostic (German). Multi: direct EN+DE.}\\
\multicolumn{5}{@{}l@{}}{\scriptsize $^\dag$Wide CI. $^\ddag$Whisper-S two-stage = layer-wise analysis checkpoint.}
\end{tabular}
\end{table}

\noindent\textbf{Language probe} classifies English medical vs. German medical speech. Because the two language subsets come from different corpora, the contrast may also reflect speaker and acoustic-condition differences.

These confounds should be considered when interpreting the near-ceiling probe F1 scores reported in Sec.~VI.C.

\noindent\textbf{WER probe} classifies low- vs. high-WER utterances. WER is computed per checkpoint using greedy decoding on 100 English test clips and labelled by rank-based tertile split: bottom-third (low-WER, label=0) vs.\ top-third (high-WER, label=1), middle third excluded (n=33 per class, applied identically across checkpoints).

For each encoder layer, variable-length hidden-state sequences were converted into fixed-dimensional representations by mean-pooling across the temporal dimension. Each task uses Logistic Regression, Linear SVC, and a one-hidden-layer MLP trained on frozen encoder features with stratified 5-fold cross-validation\cite{alain2018understandingintermediatelayersusing}. Panel-mean macro-F1 is the primary metric. Features are MinMax-scaled and reduced to 50 principal components to limit overfitting under small sample sizes. To assess robustness, representation drift and probing analyses are repeated over five bootstrap seeds: 42, 123, 456, 789, and 2024. Layer-wise differences are tested using Kruskal--Wallis tests with Bonferroni and FDR-BH correction\cite{888cd474-50a6-33fd-a789-415b80e67e78}. Within each fold, MinMax scaling and PCA reduction to 50 components are fitted only on the training split and then applied to the held-out validation split, preventing validation-fold information from entering preprocessing.

\section{Experiments \& Results}
This section reports ASR performance and representation-level analyses. We first compare zero-shot and fine-tuned Whisper models with Wav2Vec2 baselines, then analyse Whisper-Small encoder representations to study how medical and multilingual adaptation affect layer-wise information.

\subsection{ASR Performance: Zero-Shot and Fine-Tuned}

Tab.~\ref{tab:asr_main} reports zero-shot and fine-tuned MedASR WER results across all models. Among zero-shot Whisper models, Whisper-Large-v3 performs best on all subsets (English 16.50\%, German 56.51\%, combined 35.52\%). The Wav2Vec2 CTC baselines show poor out-of-domain transfer, with English WER above 92\% and German XLSR-53 reaching 76.00\%, confirming the need for domain adaptation in this medical setting.

Fine-tuning changes the model ranking. English-only fine-tuning gives the best English result with Whisper-Medium (7.72\% WER), but German transfer without German training is inconsistent: Whisper-Small degrades on German, whereas Whisper-Medium improves relative to its zero-shot baseline. The German-only diagnostic shows useful within-corpus adaptation signal, with Whisper-Large-v3 reaching 44.96\% WER and Whisper-Medium 45.94\%. Since this setting uses only 86 German training utterances, we do not interpret it as robust German generalisation.

For multilingual training, two-stage EN$\rightarrow$EN+DE adaptation improves over zero-shot for all Whisper sizes, with Whisper-Small giving the lowest two-stage combined WER (31.33\%). Direct EN+DE fine-tuning gives the best overall combined result, with Whisper-Medium reaching 26.30\% combined WER and 46.72\% German WER. This may reflect the strong EN--DE data imbalance (997 vs. 86 training utterances): direct EN+DE training learns a shared multilingual representation from the start, whereas continued EN$\rightarrow$EN+DE fine-tuning starts from an English-specialised checkpoint. Overall, model scale alone does not determine adapted MedASR performance; the strongest model depends on whether the objective is zero-shot robustness, English adaptation, German within-corpus adaptation, or combined EN+DE performance.

Because the layer-wise analysis below is conducted on the two-stage Whisper-Small trajectory, the representation-level findings should be interpreted as explaining that specific adaptation path; comparable analyses of direct EN+DE and German-only encoders are left for future work.

\textbf{Qualitative error audit.}
To complement aggregate WER, Tab.~\ref{tab:error_audit} shows representative examples from three checkpoints: zero-shot Whisper-Small (ZS-Small), two-stage Whisper-Small EN$\rightarrow$EN+DE (S2-Small, the layer-wise model), and German-only Whisper-Large-v3 (DE-Large, 44.96\% German WER). English examples compare ZS-Small and S2-Small; DE-Large is included for German rows only, where it achieves the best German WER.

\begin{table}[!t]
\caption{Representative transcription errors. ZS-Small / S2-Small = Whisper-Small zero-shot / two-stage EN$\rightarrow$EN+DE; DE-Large = German-only Whisper-Large-v3. Italics mark the contested term.}
\label{tab:error_audit}
\centering
\scriptsize
\renewcommand{\arraystretch}{1.12}
\setlength{\tabcolsep}{3pt}
\resizebox{\linewidth}{!}{%
\begin{tabular}{@{}p{0.19\linewidth}p{0.77\linewidth}@{}}
\toprule
\textbf{Type} & \textbf{Utterances} \\
\midrule
EN symptom term &
REF: stomach pain and \emph{bloating}; ZS: stomach pain and \emph{rotting}; S2: stomach pain and \emph{bloating}. \\[2pt]
EN clinical term &
REF: could I have a \emph{concussion}; ZS: could I have a \emph{competition}; S2: could I have a \emph{concussion}. \\
\midrule
DE imaging term &
REF: \emph{R\"{o}ntgenthorax}, \emph{Pneu};
ZS: \emph{R\"{o}ntgen Torax}, \emph{Pneuer};
S2: r\"{o}ntgen torax, \emph{panor};
DE-Large: r\"{o}ntgen thorax, \emph{pneu}. \\[2pt]
DE anatomy terms &
REF: \emph{Klavikola}, \emph{Vene}, \emph{Terumo-Nadel};
ZS: \emph{Gravikula}, \emph{Wene}, Terumo;
S2: \emph{gravikulare}, vene, \emph{pteromunadel};
DE-Large: \emph{klavikula}, vene, terumonadel. \\[2pt]
DE hallucination &
REF: die \emph{Prostater} dr\"{u}ben;
ZS/S2: \emph{Brotzeit} verdr\"{u}cken/verdr\"{u}ben;
DE-Large: die \emph{Prostata} dr\"{u}ben. \\
\bottomrule
\end{tabular}}
\end{table}

\subsection{Representation Drift}

\begin{figure}[!htpb]
    \centering
    \includegraphics[width=1\linewidth]{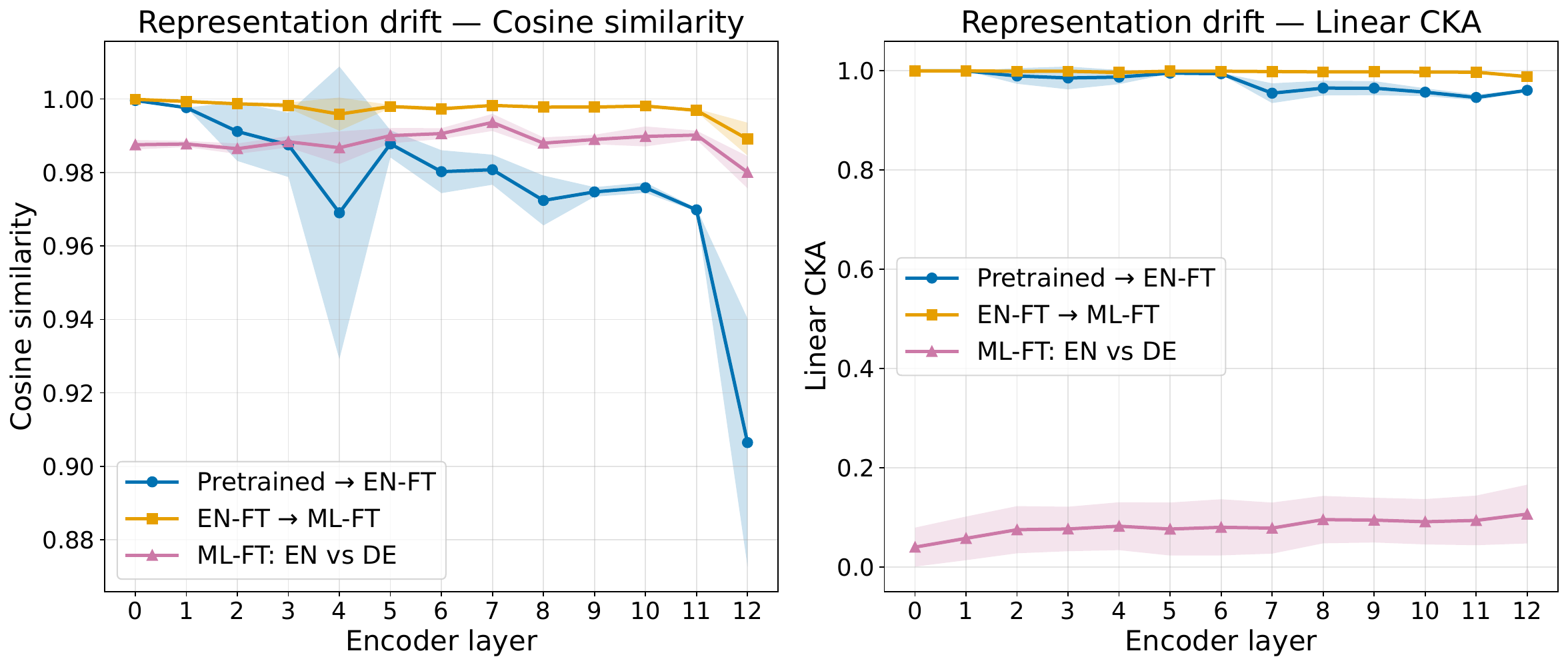}

    \caption{Bootstrap-averaged representation drift across encoder layers (L0--L12) over five seeds; shading denotes $\pm$1 std. Cosine similarity is shown on the left and Linear CKA on the right. Pretrained$\to$EN-FT produces the largest upper-layer shift, whereas EN-FT$\to$ML-FT remains close to the adapted encoder. ML-FT EN vs.\ DE shows high cosine similarity but low Linear CKA, indicating centroid alignment with language-sensitive geometry.}
    \label{fig:drift_bootstrap}
\end{figure}

The drift analysis in Fig.~\ref{fig:drift_bootstrap} reveals a clear asymmetry between the two adaptation stages. Pretrained$\rightarrow$EN-FT produces the largest representational shift, most pronounced in upper layers (L8--L12), indicating that English medical fine-tuning drives the main encoder reorganisation. In contrast, EN-FT$\rightarrow$ML-FT shows substantially smaller drift, suggesting that multilingual continuation largely preserves the English-adapted representation space. 

The ML-FT English vs. German comparison yields high cosine similarity (0.980--0.994) but low Linear CKA (0.040--0.107), indicating centroid-level cross-lingual alignment while the full representation geometry remains language-sensitive. Tab.~\ref{tab:drift_summary} summarises the bootstrap-averaged per-layer ranges.

\begin{table}[!htpb]
\caption{Representation drift summary over five bootstrap seeds. Values report the per-layer range across L0--L12.}
\label{tab:drift_summary}
\centering
\scriptsize
\setlength{\tabcolsep}{18pt}
\resizebox{0.88\linewidth}{!}{
\begin{tabular}{@{}lcc@{}}
\toprule
\textbf{Drift pair} & \textbf{Cosine range} & \textbf{Lin.\ CKA range} \\
\midrule
Pretrained $\to$ EN-FT
& 0.906--1.000
& 0.946--1.000 \\

EN-FT $\to$ ML-FT
& 0.989--1.000
& 0.988--1.000 \\

ML-FT: EN vs.\ DE
& 0.980--0.994
& 0.040--0.107 \\
\bottomrule
\end{tabular}}
\end{table}

\subsection{Probing Results}
\begin{figure*}[!t]
    \centering
    \includegraphics[width=0.98\textwidth]{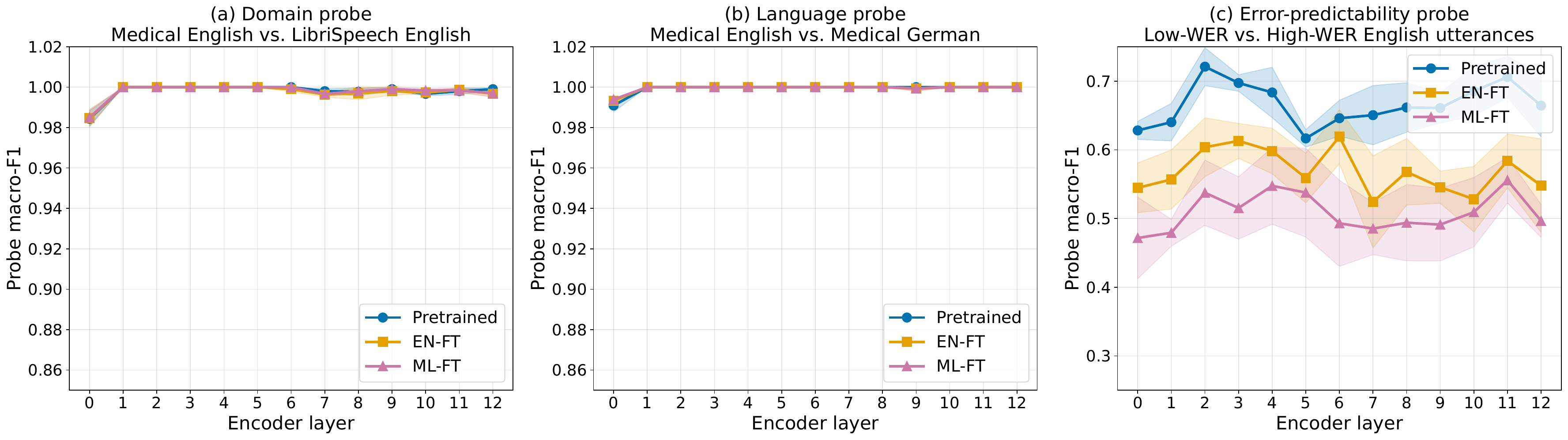}
    \caption{Layer-wise probe macro-F1 across encoder layers (L0--L12) for the three probing tasks, bootstrap-aggregated over five seeds (shading: $\pm$1 std). (a)~Domain probe (medical English vs.\ LibriSpeech) and (b)~Language probe (English medical vs.\ German medical) remain near ceiling across all checkpoints and layers, so peak-layer claims are not meaningful here. (c) Error-predictability probe uses rank-based low-/high-WER tertile labels. The pretrained encoder carries the strongest linearly recoverable error signal at the best layer; this signal weakens after English fine-tuning and multilingual continuation while decoder WER improves on the same 100 clips (19.87\%$\rightarrow$13.61\%$\rightarrow$12.48\%).} 
    \label{fig:probe_lines}
\end{figure*}

\textbf{Domain probe.} 
Fig.~\ref{fig:probe_lines}(a) shows that domain information remains highly separable across all checkpoints and layers, with panel-mean macro-F1 at or near ceiling ($\geq 0.984$; see Tab.~\ref{tab:probe_summary}). Because many layers achieve nearly identical F1, exact peak-layer claims are not meaningful. The main finding is that medical vs. general speech remains consistently recoverable from frozen encoder representations across adaptation stages. For ML-FT, silhouette scores decrease from 0.33 at L0 to 0.11 at L11, suggesting that deeper-layer domain representations remain classifiable but less geometrically compact.

\textbf{Language Probe.} Fig.~\ref{fig:probe_lines}(b) and Tab.~\ref{tab:probe_summary} show that language-probe F1 remains near ceiling under bootstrap aggregation. Language identity is almost perfectly separable across checkpoints and layers, with bootstrap mean F1 $\geq 0.990$ from L1 onward across all checkpoints. This indicates that Whisper's multilingual pretraining encodes robust language-discriminative features that are preserved through both fine-tuning stages. Since performance is at ceiling across nearly all depths, the result is best interpreted as evidence of preserved language-discriminative structure rather than a meaningful single-layer peak.

\textbf{WER/Error-Predictability Probe.} The WER/error-predictability probe in Fig.~\ref{fig:probe_lines}(c) and Tab.~\ref{tab:probe_summary} shows an overall decrease in linearly recoverable error information after adaptation, although the layer-wise curves fluctuate across encoder depth. Using rank-based low-/high-WER tertile labels, the best-layer panel-mean macro-F1 drops from $0.721 \pm 0.028$ at L2 for the Pretrained encoder (mean across layers: 0.666), to $0.619 \pm 0.039$ at L6 after EN-FT (mean: 0.569), and to $0.556 \pm 0.033$ at L11 after ML-FT (mean: 0.509, close to binary-chance level). In parallel, decoder WER on the same 100 English clips decreases from 19.87\% to 13.61\% and 12.48\%. Thus, as decoding improves, residual errors become less linearly predictable from frozen encoder activations, suggesting that adaptation reorganises failure modes that were more separable in the pretrained representation space.

Kruskal--Wallis tests pooling 13 layers over five bootstrap seeds 
($n{=}65$ observations per test) show significant layer-wise variation 
for all domain and language probes after Bonferroni and FDR-BH 
correction (all adjusted $q < 10^{-4}$). 
For the WER probe, only the pretrained checkpoint remains significant 
after correction ($p = 1.47\times10^{-3}$; Bonferroni $q = 1.32\times10^{-2}$; 
FDR-BH $q = 1.89\times10^{-3}$), whereas EN-FT ($p = 0.051$) and 
ML-FT ($p = 0.213$) are not significant under either correction. 
Domain and language information thus remain robustly layer-dependent, 
while error-predictive information becomes weaker and no longer 
significantly layer-localised after fine-tuning.

\begin{table}[!htpb]
\caption{Probe summary over five bootstrap seeds. Domain and Language report the minimum macro-F1 across layers; WER F1 reports the best-layer low-/high-WER probe result under a rank-based tertile split ($n{=}33$ per class).}
\label{tab:probe_summary}
\centering
\scriptsize
\renewcommand{\arraystretch}{1.12}
\setlength{\tabcolsep}{3pt}
\resizebox{0.95\linewidth}{!}{
\begin{tabular}{@{}lcccc@{}}
\toprule
\textbf{Checkpoint} &
\textbf{Domain F1} &
\textbf{Lang. F1} &
\textbf{WER F1} &
\textbf{Dec. WER} \\
&
\textbf{min} &
\textbf{min} &
\textbf{best layer} &
\textbf{(\%)} \\
\midrule
Pretrained & $\geq 0.984$ & $\geq 0.990$ & $0.721 \pm 0.028$ (L2)  & 19.87 \\
EN-FT      & $\geq 0.984$ & $\geq 0.993$ & $0.619 \pm 0.039$ (L6)  & 13.61 \\
ML-FT      & $\geq 0.985$ & $\geq 0.993$ & $0.556 \pm 0.033$ (L11) & 12.48 \\
\bottomrule
\multicolumn{5}{@{}p{0.95\columnwidth}@{}}{\scriptsize
Dec.\ WER is computed on the same 100 English utterances used to label the WER probe (greedy decoding) and is not layer-wise.}
\end{tabular}}
\end{table}

\section{Discussion \& Conclusion}

This work investigates medical ASR using multi-size Whisper
evaluation, English--German adaptation strategies, and layer-wise
encoder analysis. Fine-tuning substantially improves MedASR over
zero-shot decoding, but model scale alone does not determine the
best adapted model under limited clinical data: Whisper-Medium is
strongest under English-only and direct EN+DE fine-tuning, while
Whisper-Large-v3 leads under the German-only diagnostic, which we
interpret cautiously as within-corpus adaptation on 86 single-speaker training utterances.

The layer-wise analysis yields three messages.
\textbf{(i) Medical adaptation dominates representational change.}
English medical fine-tuning produces the largest encoder drift,
concentrated in upper layers.
\textbf{(ii) Multilingual continuation preserves learned
representations.} EN-FT$\rightarrow$ML-FT drift is substantially
smaller, and the domain and language probe contrasts remain highly
separable across layers; their near-ceiling separability partly
reflects corpus, speaker, and acoustic-condition confounds
(Sec.~V-C), so exact peak-layer claims should be
treated cautiously.
\textbf{(iii) Improved ASR does not imply stronger linear error
signals.} The WER probe weakens after each adaptation step even as
decoder WER improves, suggesting that fine-tuning reduces simple
linearly recoverable failure cues in the encoder.

These findings complement standard WER evaluation and suggest
several directions. The persistence of domain and language
information across layers motivates parameter-efficient or selective fine-tuning. The weakening of WER-predictive signals suggests a role for difficulty-aware data selection using pretrained encoder representations. The strong separability of the English--German medical contrast motivates further study of shared multilingual encoder architectures for clinical ASR.

\textit{Limitations.} The medical corpora are small, particularly
the single-procedure German subset (86 train / 38 test), and probing uses 100--138 utterances per task; the layer-wise analysis covers only Whisper-Small. Fine-grained layer-specific, cross-speaker, and cross-scale conclusions should therefore be treated cautiously. Future work should validate these findings on larger multilingual clinical corpora and extend the layer-wise analysis to larger Whisper variants, direct EN+DE, German-only, and parameter-efficient adaptation strategies.

\section{Generative AI Use Disclosure}
Generative artificial intelligence tools were used to assist with language editing, clarity of presentation, and code drafting/debugging. All research ideas, methodology, experiments, analyses, and interpretations were conceived, verified, and carried out by the authors, who take full responsibility for the originality, validity, and integrity of the work.

\bibliographystyle{IEEEtran}
\bibliography{references}

\end{document}